\documentclass[11pt,a4paper]{article}
\usepackage[hyperref]{acl2021}
\usepackage{times}
\usepackage{latexsym}

\usepackage{microtype}
\usepackage{graphicx}
\usepackage{booktabs}
\usepackage{multirow}
\usepackage{amsthm,amsmath,amssymb}
\usepackage{mathrsfs}
\usepackage{url}
\usepackage[linesnumbered,ruled,vlined,algo2e]{algorithm2e}
\usepackage{arydshln}
\usepackage{xspace}
\usepackage{CJK}
\usepackage{relsize}
\usepackage{subfigure}

\definecolor{electricviolet}{rgb}{0.56, 0.0, 1.0}

\aclfinalcopy 

\title{Progressive Multi-Granularity Training for \\Non-Autoregressive Translation}
  
\author{
Liang Ding\thanks{~~~Liang Ding and Longyue Wang contributed equally to this work. Work was done when Liang Ding and Xuebo Liu were interning at Tencent AI Lab.}\\
The University of Sydney\\
\normalsize \textsf{ldin3097@sydnye.edu.au}\\
\And
Longyue Wang$^*$\\
Tencent AI Lab\\
\normalsize \textsf{vinnylywang@tencent.com}\\
\And
Xuebo Liu\\
University of Macau\\
\normalsize \textsf{nlp2ct.xuebo@gmail.com}\\
\AND
Derek F. Wong\\
University of Macau\\
\normalsize \textsf{derekfw@um.edu.com}\\
\And
Dacheng Tao\\
JD Explore Academy, JD.com\\
\normalsize \textsf{dacheng.tao@gmail.com}\\
\And
Zhaopeng Tu\\
Tencent AI Lab\\
\normalsize \textsf{zptu@tencent.com}\\
}

\date{}

\begin{document}
\maketitle
\begin{abstract}
Non-autoregressive translation (NAT) significantly accelerates the inference process via predicting the entire target sequence. However, recent studies show that NAT is weak at learning high-mode of knowledge such as one-to-many translations. We argue that modes can be divided into various granularities which can be learned from easy to hard. In this study, we empirically show that NAT models are prone to learn fine-grained lower-mode knowledge, such as words and phrases, compared with sentences. Based on this observation, we propose progressive multi-granularity training for NAT. More specifically, to make the most of the training data, we break down the sentence-level examples into three types, i.e. words, phrases, sentences, and with the training goes, we progressively increase the granularities. Experiments on Romanian-English, English-German, Chinese-English and Japanese-English demonstrate that our approach improves the phrase translation accuracy and model reordering ability, therefore resulting in better translation quality against strong NAT baselines. Also, we show that more deterministic fine-grained knowledge can further enhance performance. 
\end{abstract}

\section{Introduction}
\label{sec:intro}
Non-autoregressive translation~(\citealp[NAT,][]{NAT}) has been proposed to improve the decoding efficiency by predicting all tokens independently and simultaneously. 
Different from autoregressive translation~(\citealp[AT,][]{transformer}) models that generate each target word conditioned on previously generated ones,
NAT models suffer from the multimodality problem (i.e. multiple translations for a single input), in which the conditional independence assumption prevents a model from properly capturing the highly multimodal distribution of target translations. 
To reduce the modes of training data, sequence-level knowledge distillation (KD)~\cite{kim2016sequence} is widely employed via replacing their original target samples with sentences generated from an AT teacher~\cite{NAT,zhou2019understanding, ren2020astudy}. 

\begin{table}[t]
\centering
\begin{tabular}{lrrrrr}
\toprule
\multirow{2}{*}{\bf Granular.} & \multirow{2}{*}{\bf AT}  & \multicolumn{4}{c}{\bf NAT}\\
\cmidrule(lr){3-6}
 & & Raw & $\bigtriangleup$ & KD  & $\bigtriangleup$\\
\midrule
\textsc{Word}   &   59.8 & 57.1 & -2.7 & 59.0 & -0.8\\
\textsc{Phrase}   &    36.0 & 31.7 & -4.3 & 34.2 & -1.8\\
\textsc{Sentence}   &    29.2 & 24.5 & -4.7 & 27.0 & -2.2\\
\bottomrule
\end{tabular}
\caption{Translation performance at different granularity on the WMT14 English$\Rightarrow$German dataset. ``$\bigtriangleup$'' indicates the performance gap between the NAT and AT.}
\label{tab:granular-eva}
\end{table}

Although KD reduces the learning difficulty for NAT, there are still complicated word orders and structures~\cite{gell2011origin} in the synthetic sentences, making the NAT performance sub-optimal. 
To answer this challenge, \newcite{saharia2020non,Ran2021GuidingNN} propose to lowers the bilingual modeling difficulties under the \textit{monotonicity assumption}, where bilingual sentences are in the same word order. However, they make extensive modifications to model structures or objectives, limiting the applicability of their methods to a boarder range of tasks and languages.

Accordingly, we turn to break down the sentence-level high modes into finer granularities, i.e. bilingual words and phrases, where we assume that finer granularities are easy to be learned by NAT.
As shown in Table~\ref{tab:granular-eva}, we analyzed the translation accuracy at three linguistic levels (i.e. word, phrase and sentence) and found that although KD brings promising improvements at three granularities, there are still some gaps with AT teacher. Also, we showed that finer granularities are easier to be learned, that is, accuracy gap ``$\Delta$'' of \textsc{word} is small than that of \textsc{phrase}, and \textsc{sentence} (0.8$<$1.8$<$2.2). Thus,  
we propose a simple and effective training strategy to enhance the ability to handle the sentence-level high modes.
More specifically, we generate bilingual lexicons from parallel data by leveraging word alignment and phrase extraction in statistical machine translation (\citealp[SMT,][]{zens2002phrase}). Then we guide the NAT model to progressively learn the bilingual knowledge from low to high granularity. 
Experimental results on four commonly-cited translation benchmarks show that our proposed \textsc{progressive multi-granularity} (PMG) training strategy consistently improves the translation performance. 
The main contributions are:
\begin{itemize}
    \item Our study reveals that NAT is better at learning fine-grained knowledge. Training with sentences merely may be sub-optimal.
    \item We propose PMG training to encourage NAT models to learn from easy to hard. The fine-grained knowledge distilled by SMT will be dynamically transferred during training.
    \item Experiments across language pairs and model structures show the effectiveness and universality of PMG training.
\end{itemize}

\begin{CJK}{UTF8}{gbsn}
\begin{table}[t]
\centering
\renewcommand\arraystretch{1.05}
\begin{tabular}{c|p{2.5cm}|p{3.2cm}}
\hline
& {\bf Source}   & {\bf Targets}\\
\hline
\multirow{3}{*}{\rotatebox[origin=c]{90}{{\bf Word}}} & \multirow{3}{*}{bank} &  \small 银行 \\
& & \small 岸\\
& & \small 储库\\
\hline
\multirow{3}{*}{\rotatebox[origin=c]{90}{{\bf Phrase}}} & \multirow{3}{*}{\parbox{1.5cm}{hollow structural}} & \small 中空 结构\\
& & \small 空心 的 结构\\
& & \small 镂空 结构\\
\hline
\multirow{3}{*}{\rotatebox[origin=c]{90}{{\bf Sentence}}} & \multirow{3}{*}{\parbox{2.5cm}{He is very good at English.}} & \small 他 英文 很 好。\\
& & \small 他 非常 擅长 英语 。\\
& & \small 他的 英语 水平 很 高 。\\
\hline
\end{tabular}
\caption{Examples of different translation granularities.}
\label{tab:granularity-example}
\end{table}
\end{CJK}

\section{Methodology}

\subsection{Motivation}
We investigated theories in second-language acquisition: one usually learns a foreign language from word-to-word translation to sentence-to-sentence translation, 
namely from local to global~\cite{onnis2008learn}.
Bilingual knowledge is at the core of adequacy modeling~\cite{Tu:2016:ACL}, which is a major weakness of the NAT models due to the lacks of autoregressive factorization. 
\begin{CJK}{UTF8}{gbsn}
Table~\ref{tab:granularity-example} demonstrates the English$\Rightarrow$Chinese multimodality at different granularities (i.e. word, phrase, sentence levels). As seen, the sentence-level consists of various kinds of modes, including word alignment (``English'' vs. ``{\small 英语}''/``{\small 英文}''), phrase translation (``be good at'' vs. ``...{\small 非常 擅长}...''/``...{\small 水平 很高}''), and even reordering (``{\small 英语}'' can be subject or object). However, phrase-level modes are less complex with similar structure and word-level modes are simple with token-to-token mapping.
Generally, the lower level of bilingual knowledge, the easier for NAT to learn. This example explains why the sentence level performance gaps between NAT and AT are significant than that of word and phrase in Table~\ref{tab:granular-eva}.
Based on the above evidence, it is natural to suspect that the existing sentence-level NAT training is sub-optimal.
\end{CJK}

\subsection{Fine-grained Bilingual Knowledge}
\label{ssec:bilingual}
Phrase table is an essential component of SMT systems, which records the correspondence between bilingual lexicons~\cite{koehnstatistical}.
For each training example in the original training set, we sample its all possible inter-sentence bi-lingual phrases from the phrase table that obtained with phrase-based statistical machine translation (PBSMT) model~\cite{koehn2003statistical}.
The GIZA++~\cite{och03:asc} was employed to build word alignments for the training datasets.
We leave the exploitation of more advanced forms bilingual knowledge such as syntax rules~\cite{liu2006tree} and discontinuous phrases~\cite{galley2010accurate} for future work. 
\begin{CJK}{UTF8}{gbsn}
 Take the sentence pair in Table~\ref{tab:granularity-example} for example, we can obtain the bi-lingual En-Zh phrase pairs ``\textit{very good} $|||$ \textit{\small 很好}'', ``\textit{good at English} $|||$ \textit{\small 擅长英语}'' from original sentence pair, informing the NAT model the explicit phrase boundaries. 
\end{CJK}

\begin{table*}[t]
    \centering
    \scalebox{1}{
    \begin{tabular}{lrllll}
    \toprule
     \multirow{2}{*}{\textbf{Models}} &  \multirow{2}{*}{\bf Speed} & \multicolumn{4}{c}{\bf BLEU} \\
     \cmidrule(lr){3-6}
     &&{\bf Ro-En} & {\bf En-De} & {\bf Zh-En} & {\bf Ja-En} \\
    \midrule
    \multicolumn{6}{c}{\emph{AT Models}}\\
    Transformer-\textsc{Base} (Ro-En Teacher)  &  1.0$\times$ & 34.1  & 27.3 & 24.4 & 29.2\\
    Transformer-\textsc{Big} (En-De / Zh-En / Ja-En Teacher)  & 0.8$\times$  &  n/a  &  29.2  & 25.3   & 29.8\\
    \midrule
    \multicolumn{6}{c}{\emph{Existing NAT Models}}\\
    NAT~\cite{NAT}   &   2.4$\times$ &31.4 &19.2  &n/a & n/a\\
    Iterative NAT~\cite{lee2018deterministic} & 2.0$\times$  &30.2 &21.6 &n/a & n/a\\
    DisCo~\cite{kasai2020parallel} & 3.2$\times$  &33.3& 26.8 &n/a & n/a\\
    Levenshtein~\cite{gu2019levenshtein} & 3.5$\times$  &33.3 &   27.3& n/a & n/a\\
    Mask-Predict~\cite{ghazvininejad2019mask}  &   1.5$\times$ & 33.3 & 27.0 & 23.2 & n/a\\
    {Context-aware NAT}~\citep{ding2020localness} & {1.5$\times$}   & 33.2 & 27.5 & 24.6 & 29.4\\
    \midrule
    \multicolumn{6}{c}{\emph{Our NAT Models}}\\
    {\bf Levenshtein}~\citep{gu2019levenshtein}              & \multirow{2}*{3.5$\times$}  &33.2 & 27.4& 24.4 & 29.1\\
    ~~~{\bf +PMG Training}  &  &  33.8$^\dagger$  &27.8 & 25.0$^\dagger$ &  29.6\\
     {\bf Mask-Predict}~\cite{ghazvininejad2019mask}      &  \multirow{2}*{1.5$\times$} & 33.3 & 27.0  & 24.0 & 28.9\\
    ~~~{\bf +PMG Training} &  & 33.7 &  27.6$^\dagger$  & 24.5 & 29.5$^\dagger$\\
    \bottomrule
    \end{tabular}}
    \caption{Comparison with previous work on WMT16 Ro-En, WMT14 En-De, WMT17 Zh-En and WAT17 Ja-En datasets. ``$^\dagger$'' indicates that the proposed method was significantly better than baseline at significance level {p\textless 0.05}.}
    \label{tab:main-results}
\end{table*}

\subsection{Progressive Multi-Granularity Training}
\label{ssec:PMG}
We present an extremely simple progressive multi-granularity (PMG) training fashion. 
Concretely, we progressively schedule the PMG: learn from ``low'' to ``high'' granularity, i.e. word$\rightarrow$phrase$\rightarrow$sentence. And we empirically set the training steps for each training stage. 
Our work can be seen as a typical determinism-based curriculum learning (CL)~\cite{bengio2009curriculum} method, where the finer granularities are more deterministic than sentences. Thus we compare with typical CL works~\cite{zhang2019curriculum,platanios2019competence} in Section~\ref{ssec:main-results}.

\section{Experiment}
\subsection{Setup}
\paragraph{Data}
Experiments were conducted on four widely-used translation datasets: WMT14 English-German (En-De), WMT16 Romanian-English (Ro-En), WMT17 Chinese-English (Zh-En) and WAT17 Japanese-English (Ja-En), which consist of 4.5M, 0.6M, 20M and 2M sentence pairs, respectively. It is worthy noting that Ro-En, En-De and Zh-En are low-, medium- and high- resource language pairs, and Ja-En is word order divergent language direction. We use the same validation and test datasets with previous works for fair comparison. To avoid unknown works, we preprocessed data via byte-pair encoding (BPE)~\cite{Sennrich:BPE} with 32K merge operations. 
We evaluated the translation quality with \textsc{BLEU}~\cite{papineni2002bleu} with statistical significance test~\cite{collins2005clause}.
For fine-grained bilingual knowledge, e.g. word alignment and phrase table, to ensure the source to target mapping more deterministic, we set 0.05 as the probability threshold. {Taking WMT14 En-De for example, there are 3M words and 156M phrases in the original phrase table extracted by SMT methodology. We then filter the items whose translation probability is lower than 0.05 and obtain 0.3M words and 56.5M phrases as the final data.}

\begin{figure*}[t]
    \centering
    \subfigure[Translation Quality]{
    \includegraphics[height=0.2\textheight]{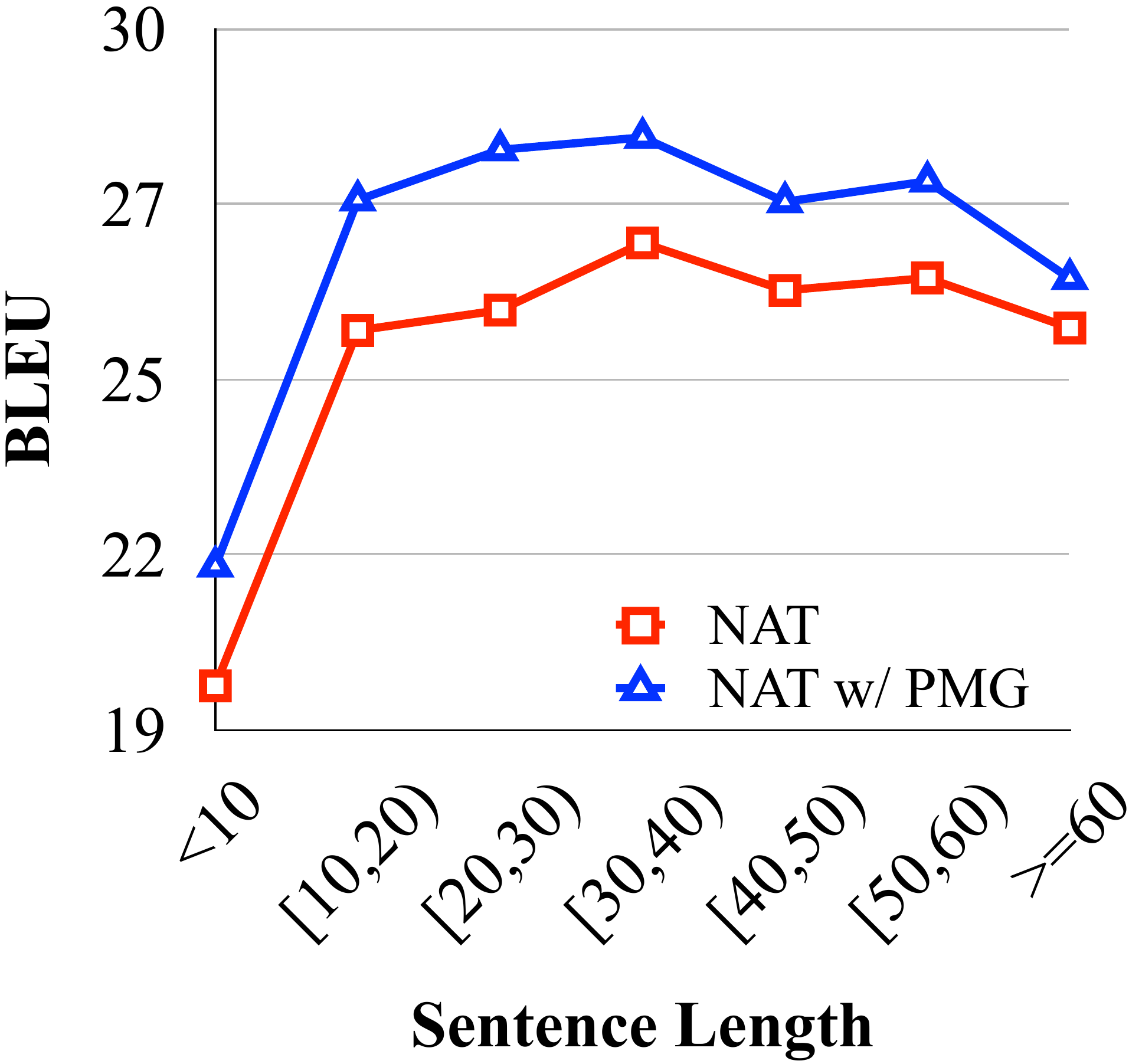}}
    \hspace{0.1\textwidth}
    \subfigure[Reordering Difference]{
    \includegraphics[height=0.2\textheight]{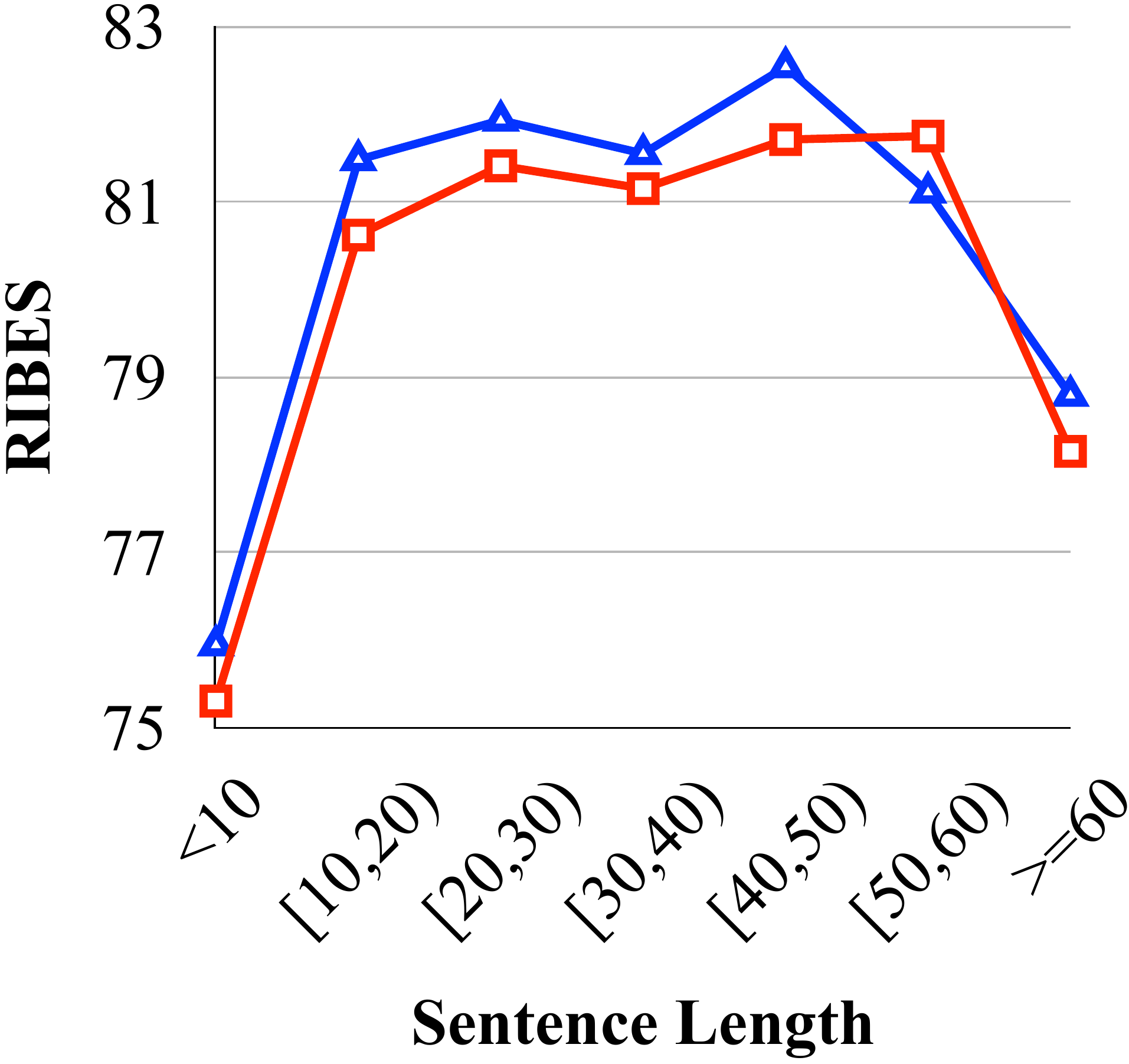}}
    \caption{Performances of our proposed approach on different length bins against the vanilla NAT model.}
    \label{fig:analyze-reorder}
\end{figure*}
\paragraph{Non-Autoregressive Models}

We validated our progressive multi-granularity training strategy on two state-of-the-art NAT model structures: 
\begin{itemize}
    \item {\em Mask-Predict} (MaskT,~\citealt{ghazvininejad2019mask}) that uses the conditional mask LM~\citep{devlin2019bert} to iteratively generate the target sequence from the masked input;
    \item {\em Levenshtein Transformer} (LevT,~\citealt{gu2019levenshtein}) that introduces three steps: {deletion}, {placeholder prediction} and {token prediction}.
\end{itemize}
For regularization, we empirically set the dropout rate as 0.2, and apply weight decay with 0.01 and label smoothing with $\epsilon$ = 0.1. We train batches of approximately 128K tokens using Adam~\citep{kingma2015adam}. The learning rate warms up to $5\times10^{-4}$ in the first 10K steps, and then decays with the inverse square-root schedule. 
{We train 50k steps on word-level data and 50k steps on phrase-level data, respectively.
And then update the remaining 200K steps for sentence-level training.}
Following the common practices~\citep{ghazvininejad2019mask,kasai2020parallel}, we evaluate the performance on an ensemble of 5 best checkpoints (ranked by validation BLEU) to avoid stochasticity. 

\paragraph{{Autoregressive Teachers}}
We closely followed previous works to apply sequence-level KD. More precisely, we trained two kinds of Transformer~\citep{transformer} models, including Transformer-\textsc{Base} and Transformer-\textsc{Big}. The main results employ \textsc{Big} for all directions except Ro-En, which is distilled by \textsc{Base}. The architectures of  Transformer-\textsc{Big} utilizes a large batch (458K tokens) training strategy.

\subsection{Experimental Results}
\label{ssec:main-results}
\paragraph{Main Results}
Table~\ref{tab:main-results} lists the results of previous competitive NAT models~\cite{NAT,kasai2020parallel,gu2019levenshtein,ghazvininejad2019mask}.
Clearly, our approach ``{+PMG Training}'' consistently improves translation performance (BLEU$\uparrow$) over four language pairs. Specifically, our PMG training strategy achieves on average +0.53 BLEU scores improvements on four language pairs upon two NAT model structures. Note that our approaches introduce no extra parameters, thus does not increase any latency (``Speed'').

\paragraph{Comparison to Curriculum Learning}
The existing CL methods can be divided into two categories, ``Discretized CL (DCL)``~\cite{zhang2019curriculum} and ``Continuous CL (CCL)``~\cite{platanios2019competence}. Sentence length is the most significant variable in our multi-granularity data, therefore we implemented discretized and continuous CL with the sentence length (source side) criteria.

Our DCL setting explicitly predefined the number of data bins, while CCL method continuously samples the shorter examples with the training progresses. For DCL, we split the training samples into a predefined number of bins (5, in our case). As for CCL, we employ their length curriculum and square root competence function. We find that on WMT14 En-De dataset with MaskT model,  DCL performs worse than KD baseline (-0.6 BLEU) while CCL outperforms KD baseline by +0.3 BLEU points. Our approach (+0.6 BLEU) is the most effective one.

\subsection{Analysis}
\label{ssec:analysis}
In this section, we conducted analytical experiments to better understand what contributes to translation performance gains. Specifically, we investigate whether the PMG 1) enhance the phrasal pattern modeling ability? 2) improve the reordering? and 3) gain better performance with higher quality fine-grained knowledge?

\paragraph{Better Phrasal Pattern Modelling}
Our method is expected to pay more attention on the bi-lingual phrases, leading to better phrase translation accuracy. To evaluate the accuracy of phrase translations, we calculate the improvement over multiple granularities of n-grams in Table~\ref{tab:analyze-phrase}, our PMG training ``NAT w/ PMG'' consistently outperforms the baseline, indicating that our proposed multi-granularity training indeed raise the ability of NAT model on capturing the phrasal patterns.

\paragraph{Better Reordering Ability}
The SMT-distilled bilingual phrasal information could intuitively inform the NAT model the bi-lingual phrasal boundaries, leading to better reordering ability. We compare the reordering ability of NAT model w/ \& w/o PMG training with RIBES\footnote{\url{http://www.kecl.ntt.co.jp/icl/lirg/ribes}}~\cite{isozaki2010automatic}, which is designed for measuring the reordering performance for distant language pairs. We categorize the test set into several bins according the sentence length and report the BLEU and RIBES scores, simultaneously in Figure~\ref{fig:analyze-reorder}. As seen, the proposed PMG training strategy could improve the translation (BLEU$\uparrow$) and reordering performance (RIBES$\uparrow$), confirming our claim. Our finding is consistent with \citet{ding2020self}, where they explicitly injected the SMT-guided alignment information into the MT models, achieving better performance.

\begin{table}[]
    \centering
    \begin{tabular}{cccccc}
        \toprule
        N-gram & 2 & 3 & 4 & 5 & 6 \\
        \midrule
        $\Delta$ BLEU & 0.5 & 0.3 & 0.3 & 0.2 & 0.2\\
        \bottomrule
    \end{tabular}
    \caption{Improvements of our proposed PMG training strategy on different N-grams against vanilla NAT.}
    \label{tab:analyze-phrase}
\end{table}

\paragraph{Effect of Fine-Grained Text Quality}
The acquired fine-grained bilingual knowledge, i.e. word alignments and phrase tables, still have extremely large volumes after filtering. Taking WMT14 En-De for example, there are over 56M phrase pairs after filtering with translation probability threshold 0.05. To make the knowledge being more deterministic, we control the quality of fine-grained text with the third party scorer -- BERTScore~\cite{zhang2020bertscore}. As illustrated in Table~\ref{tab:analyze-curriculum}, keeping the high quality bilingual knowledge (e.g. 50\%) can achieve further improvements, showing the great potential of our approach. {We will leave the exploration of high-quality bilingual knowledge for NAT as a future work.}

\section{Related Works}
\label{sec:relatwork}
\paragraph{Non-Autoregressive Translation}
There still exists a performance gap between AT teacher and its NAT student. To bridge this gap, many studies have been proposed. \citet{ghazvininejad2019mask,gu2019levenshtein,kasai2020parallel} designed novel model structures to considerably improve the NAT model capacity. \citet{wang2019non,Ran2021GuidingNN,Ding2021ICLR,Du:2021:ICML} explored to improve the model performance with additional training signals or objectives. \citet{guo2020incorporating,su2021non} delivered the knowledge from pretrained language models to the NAT models. 
Above works improve the NAT at the model level, while we improve NAT at the data level.

Most related to our work, \citet{Ding2021ACL} proposed data-level strategies, including reverse distillation and bidirectional distillation, to make the most of the parallel data. Differently, we break the sentences into fine-grained granularities to fully exploit the parallel data. Note that our model-agnostic method can be applied to any NAT structures.

\paragraph{Curriculum Learning}
Our proposed training strategy is a novel technique for NAT by exploiting curriculum learning (CL). Recent works have shown that CL can help the autoregressive translation (AT) models achieve fast convergence and better results~\cite{platanios2019competence,liu2020norm,zhan2021meta,zhou2021self}.
However, CL for non-autoregressive translation (NAT) models has not been well studied.
Among the few attempts, \citet{guo2020fine,liu2020task} respectively investigated ``parameter- and task-level'' curriculum learning approaches, while we proposed progressive multi-granularity training for NAT at ``data-level''. To the best of our knowledge, this is the first work to investigate the effects of different granularities of data on NAT models.

\begin{table}[]
    \centering
    \begin{tabular}{ccccc}
        \toprule
        Ratio & 10\% & 35\% & 50\% & 100\%\\
        \midrule
        $\Delta$ BLEU & +0.3 & +0.6 & +0.7 & +0.6\\
        \bottomrule
    \end{tabular}
    \caption{Improvement of PMG training strategy on different fine-grained data scales against vanilla NAT.}
    \label{tab:analyze-curriculum}
\end{table}

\section{Conclusion}
In this paper, we investigated the translation accuracy of different granularities in NAT, and found that the NAT models are better at dealing with fine-grained bilingual knowledge (e.g. words and phrases). Based on this finding, we proposed a simple progressive multi-granularity training strategy. Experiments show that our approach consistently and significantly improves translation performance across language pairs and model architectures. In-depth analyses indicate that our approach generates better word order and phrase patterns, outperforming typical curriculum learning methods.

\section*{Acknowledgments}
We are grateful to the anonymous reviewers and the area chair for their insightful comments and suggestions. Xuebo Liu and Derek F. Wong were supported in part by the Science and Technology Development Fund, Macau SAR (Grant No. 0101/2019/A2), and the Multi-year Research Grant from the University of Macau (Grant No. MYRG2020-00054-FST).

\bibliographystyle{acl_natbib}
\bibliography{acl2021}

\end{document}